\pdfoutput=1

\documentclass[11pt]{article}

\usepackage[preprint]{acl}

\usepackage{times}
\usepackage{latexsym}

\usepackage[T1]{fontenc}

\usepackage[utf8]{inputenc}

\usepackage{microtype}

\usepackage{inconsolata}

\usepackage{graphicx}

\usepackage{adjustbox}
\usepackage{makecell}
\usepackage{multicol}
\usepackage{multirow}
\usepackage{hhline}
\usepackage{booktabs}
\usepackage{color}
\usepackage{xcolor}
\usepackage{amssymb}
\usepackage{amsmath}
\usepackage{CJK}

\usepackage{amsmath}        
\usepackage{algorithm}      
\usepackage{algorithmic}    
\usepackage{amsmath}  
\usepackage{amssymb}   
\usepackage{wasysym}

\newcolumntype{L}[1]{>{\raggedright\arraybackslash}p{#1}}
\newcolumntype{C}[1]{>{\centering\arraybackslash}p{#1}}
\newcolumntype{R}[1]{>{\raggedleft\arraybackslash}p{#1}}

\DeclareSymbolFont{extraup}{U}{zavm}{m}{n}
\DeclareMathSymbol{\varheart}{\mathalpha}{extraup}{86}
\DeclareMathSymbol{\vardiamond}{\mathalpha}{extraup}{87}
\DeclareMathSymbol{\varclubsuit}{\mathalpha}{extraup}{88}

\DeclareSymbolFont{extraup}{U}{zavm}{m}{n}
\DeclareMathSymbol{\varspadesuit}{\mathalpha}{extraup}{83}
\DeclareMathSymbol{\varheartsuit}{\mathalpha}{extraup}{86}
\DeclareMathSymbol{\vardiamond}{\mathalpha}{extraup}{87}
\DeclareMathSymbol{\varclubsuit}{\mathalpha}{extraup}{88}

\definecolor{purple}{RGB}{128, 0, 128}

\title{Representation Decomposition for Learning Similarity and Contrastness Across Modalities for Affective Computing}

\author{
    Yuanhe Tian$^{\varheart}$, \hspace{0.1cm}
    Pengsen Cheng$^{\Delta}$, \hspace{0.1cm}
    Guoqing Jin$^{\Diamond}$, \hspace{0.1cm}
    Lei Zhang$^{\spadesuit}$, \hspace{0.1cm}
    Yan Song$^{{\spadesuit}*}$
    \\
    $^{\varheart}$University of Washington 
    \hspace{0.1cm}
    $^{\Delta}$Sichuan University \\
    $^{\Diamond}$People’s Daily Online \hspace{0.1cm}
    $^{\spadesuit}$University of Science and Technology of China \\
    $^{\varheart}$\texttt{yhtian@uw.edu} \hspace{0.1cm}
    $^{\Delta}$\texttt{chengpengsen@scu.edu.cn} \\
    $^{\Diamond}$\texttt{jinguoqing@people.cn} \hspace{0.1cm}
    $^{\spadesuit}$\texttt{leizh23@ustc.edu.cn}
    \hspace{0.1cm}
    $^{\spadesuit}$\texttt{clksong@gmail.com} 
}

\begin{document}
\maketitle

\renewcommand{\thefootnote}{\fnsymbol{footnote}}
\footnotetext[1]{Corresponding author.}
\renewcommand{\thefootnote}{\arabic{footnote}}

\begin{abstract}

Multi-modal affective computing aims to automatically recognize and interpret human attitudes from diverse data sources such as images and text, thereby enhancing human–computer interaction and emotion understanding. 
Existing approaches typically rely on unimodal analysis or straightforward fusion of cross-modal information that fail to capture complex and conflicting evidence presented across different modalities. 
In this paper, we propose a novel LLM-based approach for affective computing that explicitly deconstructs visual and textual representations into shared (modality-invariant) and modality-specific components.
Specifically, our approach firstly encodes and aligns input modalities using pre-trained multi-modal encoders, then employs a representation decomposition framework to separate common emotional content from unique cues, and finally integrates these decomposed signals via an attention mechanism to form a dynamic soft prompt for a multi-modal LLM.
Extensive experiments on three representative tasks for affective computing, namely, multi-modal aspect-based sentiment analysis, multi-modal emotion analysis, and hateful meme detection, demonstrate the effectiveness of our approach, which consistently outperforms strong baselines and state-of-the-art models.\footnote{Code released at \url{https://github.com/synlp/RD-AC}.}
\end{abstract}

\section{Introduction}

Multi-modal affective computing is an emerging field focused on identifying and interpreting emotion-related information across modalities from various sources of data (e.g., visual and textual content from news, social media, business conversations, etc.).
By leveraging information typically from images and texts, multi-modal affective computing aims to predict or classify affective states for tasks such as multi-modal aspect-based sentiment analysis (MABSA) \cite{wang2023dual,yang2024empirical}, and multi-modal emotion analysis (MEA) \cite{jia-etal-2022-beyond,zhang2023m3gat}, hateful meme detection (HMD) \cite{koutlis2023memefier,tian-etal-2024-learning-multimodal}, etc.
Such tasks are essential for various real-world applications, ranging from improving user experiences and moderating harmful or hateful content on social media to developing empathetic human–computer interaction systems \cite{zhong2024cross,sun2025multilayer}.
Thus, affective computing plays a pivotal role in creating safe online communities, providing deeper insights into emotional behaviors, and enabling intelligent systems to respond to users with increased sensitivity and contextual awareness.

\begin{figure}
    \centering
    \includegraphics[width=1\linewidth, trim=0 15 0 10]{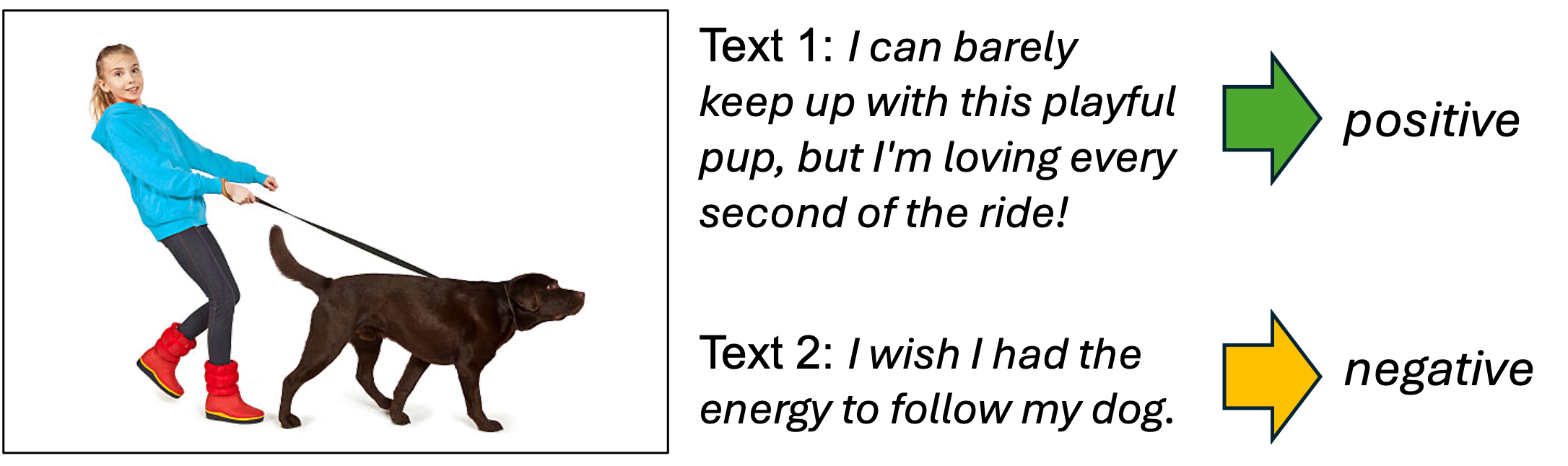}
    \caption{Examples show that whether the image matches the text may lead to different sentiments.}
    \label{fig:intro-example}
    \vspace{-1.2em}
\end{figure}

\begin{figure*}
    \centering
    \includegraphics[width=1\linewidth, trim=0 15 0 10]{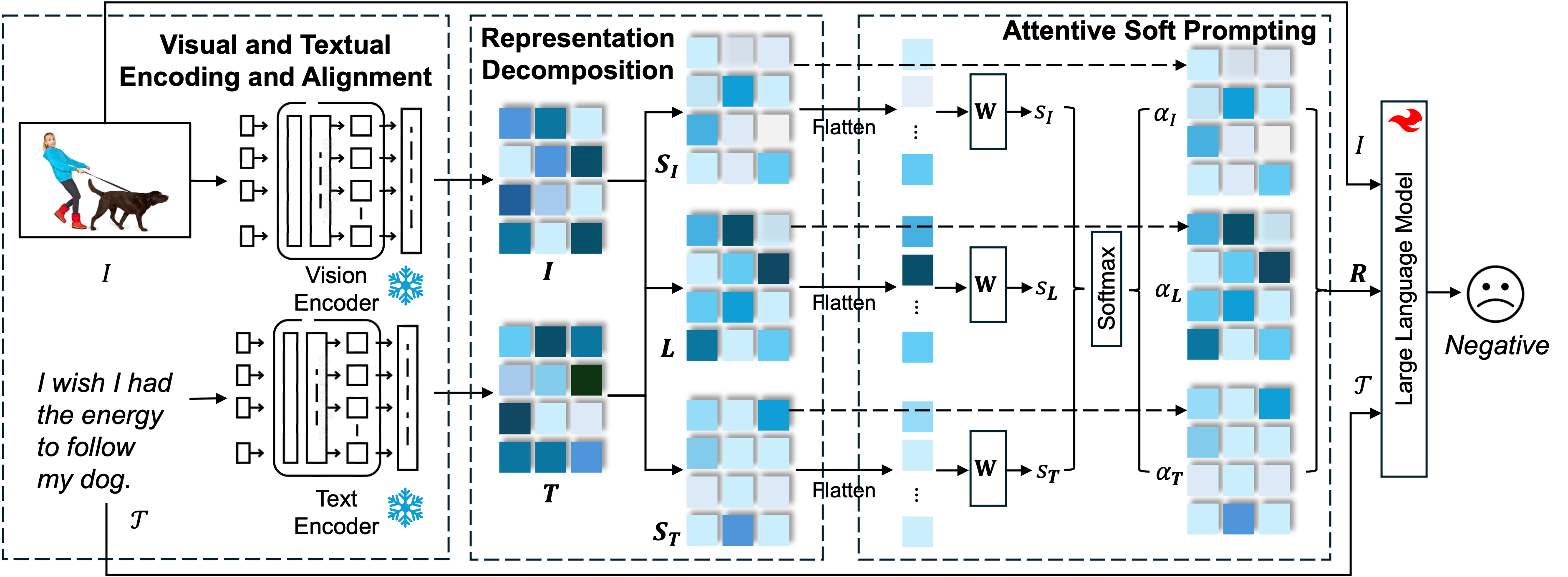}
    \caption{
    The overall architecture of our approach.
    The left part illustrates the example input image and text as well as the process of encoding and aligning them through a pre-trained multi-modal encoder.
    The middle part presents the representation decomposition, where the visual and textual representations are decomposed into a shared low-rank representation that stores the information across different modalities and two sparse matrices that contain unique information for each modality.
    The right part presents the attentive soft prompting process that employs an attention mechanism to distinguish the important information stored in the low-rank and sparse matrices, which is used as the instruction to guide the LLM for predicting corresponding results.
    }
    \label{fig:model}
    \vspace{-1em}
\end{figure*}

Conventionally, both images and texts jointly determine the final affective interpretation. 
Sometimes, the two modalities converge on a consistent meaning while in other occasions the images and texts do not carry coherent semantic information, so that creating the extra affective tendency.
For example, in Figure \ref{fig:intro-example}, the image depicts a girl being pulled forward by a brown dog, accompanied by two different text captions. The combination of the image with each caption leads to distinct sentiments. 
A model relying solely on the image might misinterpret the sentiment, while a purely text-based approach could fail to account for the visual context.
Although some attempts \cite{kiros2014multimodal,baltruvsaitis2018multimodal} to merge features from the two modalities (e.g., simple fusion or concatenation) partially capture certain correlations, they often fail to explicitly distinguish the shared, aligned portions from the more subtle or contradictory information that drives the overall affective expression.
Recent studies explore various approaches to leverage pre-trained encoders for both visual and textual modalities before fusing them for downstream predictions \cite{radford2021learning,li2022blip}, while other research incorporates external knowledge (e.g., the syntactic knowledge and keywords), data augmentation schemes, model ensemble for better results \cite{muennighoff2020vilio,qin-etal-2021-relation,zhu2022multimodal,wang2023dual,ezzameli2023emotion,koutlis2023memefier,lawan2024dualkanbaformer}. 
There are also studies \cite{singh2022flava,zhu2023minigpt,wang2024qwen2} that leverage large language models (LLMs) to benefit from the domain and task knowledge they learned from large-scale data.
However, most of these approaches pay little attention to explicitly separating the convergent aspects of each modality (i.e., the shared semantics) from the divergent or conflicting ones.
%
Thus, it is expected to have a solution that better models how the image and text jointly, yet differentially, contribute to the affective outcome.

In this paper, we propose an LLM-based approach to affective computing, which utilizes representation decomposition to explicitly separate cross-modal representations into shared and modality-specific parts, so as to capture both cooperative and conflicting information more effectively.
Specifically, we firstly encode and align the image and text into a common space, yielding the aligned visual and textual representations that are semantically comparable. 
In the next step, we decompose each aligned feature representation into two parts. 
The first part is a low-rank representation shared by both visual and textual features, capturing the common information shared across modalities. 
The second part is a sparse representation unique to either the visual or textual domain, containing their respective distinct information.
By explicitly disentangling these signals, our approach not only handles straightforward cases where the two modalities reinforce each other but also emphasizes their differences where contrasting evidences emerge and shapes the overall affective meaning.
Finally, these decomposed components are weighed and fused through an attention mechanism, where the resulting representation is fed into an LLM for the target task to predict the task label, so as to adaptively focus on shared representations when the modalities align and simultaneously exploit contrastive elements when conflicting semantics are crucial.
We evaluate our framework on benchmark datasets for representative affective computing tasks, including MABSA, MEA, and HMD.
Experiment results demonstrate consistent improvements of our approach over strong baselines and state-of-the-art models in terms of accuracy and F1-score, demonstrating its effectiveness.

\section{The Approach}
\label{sec:approach}

Based on LLMs,
our approach
is enhanced by the decomposition of aligned visual and textual representations.
Figure \ref{fig:model} illustrates the architecture of our approach.
The left-side presents the process $f_E$ to compute visual and textual representations from the input image $\mathcal{I}$ and text $\mathcal{T}$;
the middle part demonstrates the representation decomposition module $f_{RD}$ to decompose the encoded multi-modal representation into one shared and two separate representations for storing the shared and contrast information, respectively;
the right part shows the standard LLM-based decoding process to predict the class label $\widehat{y}$, where an attention mechanism $f_{Att}$ is used to weight the contribution of different information from the representation decomposition module and use the weighted representation as the soft-prompt to instruct the LLM $f_{LLM}$ to predict the final label.
The overall process of our approach is formulated as
\begin{equation}
\setlength\abovedisplayskip{5pt}
\setlength\belowdisplayskip{5pt}
    \widehat{y} = f_{LLM} (\mathcal{I}, \mathcal{T}, f_{Att} (f_{RD} (f_E(\mathcal{I}, \mathcal{T}))) )
\end{equation}
The details of the aforementioned parts are illustrated in the following text.

\subsection{Visual and Textual Encoding and Alignment}
\label{sec:3.1}

Visual and textual encoding and alignment put the image $\mathcal{I}$ and the text $\mathcal{T}$ into a unified semantic space, so as to facilitate the decomposition of their presentations.
In doing so, we leverage a pre-trained multi-modal encoding model, such as CLIP \cite{radford2021learning}, to encode them.
Specifically, for $\mathcal{I}$,
we split the image into $N$ patches $\mathcal{I} = i_1, \cdots, i_N$ (the $n$-th patch is denoted as $i_n$) and utilize a linear projection function $f_{pe}$ to map each patch $i_n$ into its patch embedding $\mathbf{e}^i_n$.
Then, we feed the patch embeddings into the vision encoder $f_{vis}$ to compute the vision representation matrix $\mathbf{I}$.
The process is formulated by
\begin{equation}
\setlength\abovedisplayskip{5pt}
\setlength\belowdisplayskip{5pt}
    \mathbf{I} = f_{vis}(\mathbf{e}^i_1 \cdots \mathbf{e}^i_N), \quad \mathbf{e}^i_n = f_{pe}(i_n)
\end{equation}
For $\mathcal{T}$, we utilize a tokenizer to split the text into tokens $\mathcal{T}=t_1 \cdots t_M$ (the $m$-th token is denoted as $t_m$) and employ an embedding layer $f_e$ to map each token $t_m$ into the token embedding $\mathbf{e}^t_m$.
We use the text encoder $f_{text}$ to encode all token embeddings and obtain the textual representation matrix $\mathbf{T}$.
The overall process is formulated as 
\begin{equation}
\setlength\abovedisplayskip{5pt}
\setlength\belowdisplayskip{5pt}
    \mathbf{T} = f_{text} (\mathbf{e}^t_1 \cdots \mathbf{e}^t_M), \quad \mathbf{e}^t_m = f_e(t_m)
\end{equation}
The visual and textual representation matrices, namely, $\mathbf{I}$ and $\mathbf{T}$, are used in the representation decomposition module for computing their shared and contrast information.

\begin{algorithm}[t!]
\caption{Algorithm for Joint Representation Decomposition}
\label{alg:admm_joint_decomp}
\small
\begin{algorithmic}[1]
\REQUIRE
\begin{itemize}
  \item $\mathbf{I}, \mathbf{T} \in \mathbb{R}^{m \times n}$: Aligned visual and textual matrices.
  \item $\lambda$: Balancing parameter for nuclear and $\ell_1$ norms.
  \item $\mu$: Penalty parameter enforcing consistency.
  \item $K$: Maximum number of iterations.
  \item $\epsilon$: Convergence tolerance.
\end{itemize}
\STATE \textbf{Initialize:}
\begin{itemize}
  \item $\mathbf{L}^{(0)} \leftarrow \mathbf{0}$, 
  $\mathbf{S}_I^{(0)}, \mathbf{S}_T^{(0)} \leftarrow \mathbf{0}$,
  $\mathbf{Z}_I^{(0)}, \mathbf{Z}_T^{(0)} \leftarrow \mathbf{0}$.
\end{itemize}
\FOR{$k = 1$ \TO $K$}
  \STATE \textbf{Update sparse components:}
  \[
    \mathbf{S}_I^{(k)} \leftarrow \mathrm{soft}\!\text{-}\mathrm{threshold}\bigl(\mathbf{I}-\mathbf{L}^{(k-1)} + \tfrac{1}{\mu}\mathbf{Z}_I^{(k-1)}, \tfrac{\lambda}{\mu}\bigr)
  \]
  \[
    \mathbf{S}_T^{(k)} \leftarrow \mathrm{soft}\!\text{-}\mathrm{threshold}\bigl(\mathbf{T}-\mathbf{L}^{(k-1)} + \tfrac{1}{\mu}\mathbf{Z}_T^{(k-1)}, \tfrac{\lambda}{\mu}\bigr)
  \]
  \STATE \textbf{Update low-rank component:}
  \[
    \mathbf{A} \leftarrow \tfrac{1}{2}\Bigl((\mathbf{I}-\mathbf{S}_I^{(k)}) + (\mathbf{T}-\mathbf{S}_T^{(k)}) + \tfrac{1}{\mu}(\mathbf{Z}_I^{(k-1)} + \mathbf{Z}_T^{(k-1)})\Bigr)
  \]
  \STATE Compute SVD: $\mathbf{A} = \mathbf{U}\,\mathbf{\Sigma}\,\mathbf{V}^\top$, then
  \[
    \mathbf{\Sigma}' \leftarrow \max(\mathbf{\Sigma} - \tfrac{1}{2\mu},0),\quad
    \mathbf{L}^{(k)} \leftarrow \mathbf{U}\,\mathbf{\Sigma}'\,\mathbf{V}^\top
  \]
  \STATE \textbf{Update multipliers:}
  \[
    \mathbf{Z}_I^{(k)} \leftarrow \mathbf{Z}_I^{(k-1)} + \mu(\mathbf{I}-\mathbf{L}^{(k)} - \mathbf{S}_I^{(k)})
  \]
  \[
    \mathbf{Z}_T^{(k)} \leftarrow \mathbf{Z}_T^{(k-1)} + \mu(\mathbf{T}-\mathbf{L}^{(k)} - \mathbf{S}_T^{(k)})
  \]
  \STATE \textbf{Check convergence:}
  \[
    \max\bigl(\|\mathbf{I}-\mathbf{L}^{(k)}-\mathbf{S}_I^{(k)}\|_F, \|\mathbf{T}-\mathbf{L}^{(k)}-\mathbf{S}_T^{(k)}\|_F\bigr) < \epsilon
  \]
  \IF{converged}
  \STATE \textbf{break}
\ENDIF
\ENDFOR
\ENSURE $\mathbf{L}^{(k)}, \mathbf{S}_I^{(k)}, \mathbf{S}_T^{(k)}$
\end{algorithmic}
\end{algorithm}

\subsection{Representation Decomposition}
\label{sec:3.2}

Multi-modal representations often contain redundancy since the inherent similarities and repeated patterns present across modalities, where
such redundancy implies that a large portion of the multi-modal data is effectively captured by a low-rank structure. 
In fact, previous studies have demonstrated that low-rank matrix recovery (LMR) is able to successfully disentangle the shared, essential information from modality-specific noise in various image analysis tasks \cite{li2024feature,tan2024research}.
This motivates this work to perform LMR to extract the essential information in the multi-modal representation.

In general,
standard LMR decomposes an input matrix \(\mathbf{X}\) into a low-rank component \(\mathbf{L}\) and a sparse error matrix \(\mathbf{S}\) via
\begin{equation} \label{eq:standard_lmr}
\setlength\abovedisplayskip{5pt}
\setlength\belowdisplayskip{5pt}
\mathbf{X} = \mathbf{L} + \mathbf{S}
\end{equation}
which is typically solved by minimizing the nuclear norm of \(\mathbf{L}\) (to promote low-rankness) and the \(\ell_1\) norm of \(\mathbf{S}\) (to encourage sparsity), which leads to the optimization problem\footnote{\textcolor{black}{This optimization problem and its solution through Lagrangian function is derived from \citet{candes2011robust}.}}
\begin{equation} \label{eq:standard_opt}
\setlength\abovedisplayskip{5pt}
\setlength\belowdisplayskip{5pt}
\min_{\mathbf{L},\,\mathbf{S}} ( \|\mathbf{L}\|_* + \lambda \|\mathbf{S}\|_1 ) \quad \text{s.t.} \quad \mathbf{X} = \mathbf{L} + \mathbf{S}
\end{equation}
where \(\lambda\) is a balancing parameter.

In our approach, we extend the standard LMR by synchronously decomposing the aligned visual and textual representation matrices to capture the shared and contrast information between modalities, namely, $\mathbf{I}$ and $\mathbf{T}$, via a shared low-rank matrix $\mathbf{L}$ and two sparse matrices $\mathbf{S}_I$ and $\mathbf{S}_T$ that satisfy
\begin{equation}
\setlength\abovedisplayskip{5pt}
\setlength\belowdisplayskip{5pt}
\mathbf{I} = \mathbf{L} + \mathbf{S}_{I}
\quad 
\text{and} 
\quad
\mathbf{T} = \mathbf{L} + \mathbf{S}_{T} 
\end{equation}
where $\mathbf{L}$ is expected to store the shared information in the multi-modalities while $\mathbf{S}_{I}$ and $\mathbf{S}_{T}$ denote modality-specific or potentially contradictory information unique to the image or text, respectively.
\textcolor{black}{
The key difference between our approach and the standard LMR is that the standard approach does not require the low-rank matrix to be the same for different modalities, whereas our approach does.
}

To obtain $\mathbf{L}$, $\mathbf{S}_{I}$, and $\mathbf{S}_{T}$, we iteratively optimize the following augmented Lagrangian function:
\begin{equation}
\begin{aligned}
& \mathcal{J}(\mathbf{L}, \mathbf{S}_I, \mathbf{S}_T, \mathbf{Z}_I, \mathbf{Z}_T) \\
& =\; \|\mathbf{L}\|_* \;+\; \lambda\Bigl(\|\mathbf{S}_I\|_1 + \|\mathbf{S}_T\|_1\Bigr) \\
& +\, \langle \mathbf{Z}_I,\, \mathbf{I} - \mathbf{L} - \mathbf{S}_I \rangle + \langle \mathbf{Z}_T,\, \mathbf{T} - \mathbf{L} - \mathbf{S}_T \rangle \\
& +\, \frac{\mu}{2}\Bigl(\|\mathbf{I} - \mathbf{L} - \mathbf{S}_I\|_F^2 + \|\mathbf{T} - \mathbf{L} - \mathbf{S}_T\|_F^2\Bigr)
\end{aligned}
\end{equation}
where \(\mathbf{Z}_I\) and \(\mathbf{Z}_T\) denote the Lagrange multipliers associated with the constraints, and \(\mu\) is the penalty parameter. 
In each iteration, we alternately update the sparse components \(\mathbf{S}_I\) and \(\mathbf{S}_T\) via element-wise soft-thresholding and update the shared low-rank matrix \(\mathbf{L}\) via singular value thresholding.
The Lagrange multipliers are subsequently adjusted to enforce the constraints and thus ensure convergence to a solution that effectively captures both the shared and modality-specific information.
The algorithm is presented in Algorithm \ref{alg:admm_joint_decomp}.
Upon the completion of optimization, the resulted $\mathbf{L}$, $\mathbf{S}_{I}$, and $\mathbf{S}_{T}$ are then prepared for next step.

\begin{table}[t]
\centering
\begin{tabular}{l|ccc}
\toprule
& \textbf{Twitter-15} & \textbf{MSED} & \textbf{HMC} \\
\midrule
\textbf{Train} & 3,179              & 6,127      & 8,500      \\
\textbf{Dev}   & 1,122              & 1,021      & 500        \\
\textbf{Test}  & 1,037              & 2,042      & 1,000      \\
\bottomrule
\end{tabular}
\caption{The statistics for the Twitter-15, MSED, and HMC datasets across the train, dev, and test splits.}
\label{tab:dataset_statistics}
\vspace{-0.5cm}
\end{table}

\begin{table*}[t]
    \centering
    \begin{tabular}{l|c c | c c | c c}
    \toprule
    & \multicolumn{2}{c|}{\textbf{Twitter-15}}
    & \multicolumn{2}{c|}{\textbf{MSED}}
    & \multicolumn{2}{c}{\textbf{HMC}} 
    \\
    & Acc & F1 & Acc & F1 & Acc & AUROC \\
    \midrule
    Qwen2-VL (2B)
    & 77.91 & 71.42 
    & 81.83 & 81.34 
    & 74.00 & 82.50 
    \\
    Qwen2-VL (2B) + LMR
    & 78.43 & 72.26
    & 82.58 & 82.16 
    & 74.60 & 83.24 
    \\
    Qwen2-VL (2B) + RD
    & 79.17 & 74.30 
    & 83.29 & 82.48 
    & 75.40 & 84.32 
    \\
    \midrule
    Qwen2-VL (2B) + RD + Att (Ours)
    & \textbf{80.96} & \textbf{76.30} 
    & \textbf{84.06} & \textbf{83.15} 
    & \textbf{76.40} & \textbf{85.04} 
    \\
    \midrule
    \midrule
    LLaVA (7B)
    & 78.04 & 71.58 
    & 82.01 & 81.67 
    & 74.40 & 82.83 
    \\
    LLaVA (7B) + LMR
    & 78.61 & 72.41
    & 82.64 & 82.22 
    & 74.80 & 83.37 
    \\
    LLaVA (7B) + RD
    & 79.30 & 74.43 
    & 83.38 & 82.53 
    & 75.60 & 84.47 
    \\
    \midrule
    
    LLaVA (7B) + RD + Att (Ours)
    & \textbf{81.11} & \textbf{76.48} 
    & \textbf{84.21} & \textbf{83.28} 
    & \textbf{76.80} & \textbf{85.25} 
    \\
    \midrule
    \midrule
    LLaVA (13B)
    & 78.16 & 71.65 
    & 82.18 & 81.72 
    & 74.80 & 83.17 
    \\
    LLaVA (13B) + LMR
    & 78.80 & 72.56
    & 82.81 & 82.31 
    & 75.20 & 83.62 
    \\
    LLaVA (13B) + RD
    & 79.52 & 74.66 
    & 83.44 & 82.62 
    & 75.80 & 84.52 
    \\
    \midrule
    LLaVA (13B) + RD + Att (Ours)
    & \textbf{81.26} & \textbf{76.52} 
    & \textbf{84.33} & \textbf{83.36} 
    & \textbf{77.00} & \textbf{85.36} 
    \\
    
    \bottomrule
    \end{tabular}
    \vspace{-0.2cm}
    \caption{
    The performance of baselines and our approach \textcolor{black}{with different metrics} on Twitter-15, MSED, and HMC for MABSA, MEA, and HMD tasks, respectively, where \textcolor{black}{Qwen2-VL (2B), LLaVA (7B), and LLaVA (13B) are} used as the foundation LLMs for this experiment.
    }
    \label{tab:overall_results}
    \vspace{-0.5cm}
\end{table*}

\subsection{Attentive Soft Prompting}
\label{sec:3.3}

Conventionally, affective computing is strongly influenced 
by how image and text correspond to each other, where in many cases they are influenced by the degree of image-text matching and other cases by the contrast between them. 
To dynamically model this process and motivated by existing studies that identify the important information by assigning weights to them \cite{vaswani2017attention,tian-etal-2020-supertagging,huang2021attention,ayetiran2023inter,tian-etal-2023-end}, we propose an attention mechanism to determine the weights for the decomposed representations \(\mathbf{L}\), \(\mathbf{S}_I\), and \(\mathbf{S}_T\), and then compute their weighted sum as a soft prompt to instruct the multi-modal LLM in predicting the label $\widehat{y}$.
Specifically, we firstly flatten \(\mathbf{L}\), \(\mathbf{S}_I\), and \(\mathbf{S}_T\) into vectors, denoted as \(\text{vec}(\mathbf{L})\), \(\text{vec}(\mathbf{S}_I)\), and \(\text{vec}(\mathbf{S}_T)\), respectively.
A fully connected layer then maps the vectors to scalar scores by
\begin{equation}
\begin{aligned}
s_L =& \mathbf{W} \cdot \text{vec}(\mathbf{L}) + b \\
s_I =& \mathbf{W} \cdot \text{vec}(\mathbf{S}_I) + b \\
s_T =& \mathbf{W} \cdot \text{vec}(\mathbf{S}_T) + b
\end{aligned}
\end{equation}
where $s_L$, $s_I$, and $s_T$ are scores for $\mathbf{L}$, $\mathbf{S}_{I}$, and $\mathbf{S}_T$, respectively, and \(\mathbf{W}\) and \(b\) are learnable parameters.\footnote{\textcolor{black}{We use the same $\mathbf{W}$ and $b$ for the following reason. Since we use a multi-modal encoder (e.g, CLIP) to compute image and text representations $\mathbf{I}$ and $\mathbf{T}$, they are in the same space. The representation decomposition process $\mathbf{I}$ and $\mathbf{T}$, and the resulting $\mathbf{L}$, $\mathbf{S}_I$, and $\mathbf{S}_T$ are in the same space. We use the same $\mathbf{W}$ and $b$ to make the resulting values also in the same space, so that they are comparable with each other.}}
The scores are then normalized using the softmax function to obtain the corresponding attention weights, namely, $\alpha_L$, $\alpha_I$, and $\alpha_T$, through
\begin{equation}
\alpha_L, \alpha_I, \alpha_T = \text{softmax} (s_L, s_I, s_T)
\end{equation}
With these weights, we compute the aggregated representation by
\begin{equation}
    \mathbf{R} = \alpha_L\,\mathbf{L} + \alpha_I\,\mathbf{S}_I + \alpha_T\,\mathbf{S}_T
\end{equation}
where $\mathbf{R}$ is employed as a soft prompt to instruct the LLM \(f_{LLM}\) in predicting final label by
\begin{equation}
    \widehat{y} = f_{LLM}(\mathcal{I}, \mathcal{T}, \mathbf{R})
\end{equation}
where
for training, $\widehat{y}$ is compared with the ground truth $y^*$ to compute the cross-entropy loss $\mathcal{L}$. 
The parameters of the LLM \(f_{LLM}\) are then optimized to minimize this loss, while the parameters of the visual-textual encoding and representation decomposition modules remain fixed.

\section{Experiment Settings}

\subsection{Tasks and Datasets}

In our experiments, we explore three representative affective computing tasks: multi-modal aspect-based sentiment analysis (MABSA), multi-modal emotion analysis (MEA), and hateful meme detection (HMD).
\textcolor{black}{
For the MABSA task, we employ the Twitter-15 dataset \cite{ijcai2019p0751}, which comprises multi-modal Twitter posts annotated with sentiment labels (positive, neutral, and negative).
We follow the standard training, development, and test splitting for it.
The MEA task is conducted using the MSED dataset \cite{jia-etal-2022-beyond}, containing
text-image pairs collected from social media. 
Each instance is manually annotated with one of six emotion labels: \textit{Anger}, \textit{Disgust}, \textit{Fear}, \textit{Happy}, \textit{Neutral}, or \textit{Sad}. 
For the HMD task, we utilize the hateful memes challenge (HMC) dataset \cite{kiela2020hateful}, 
where each sample comprises an image and its accompanying text.
This dataset is specifically curated for the detection of hateful content in multi-modal scenarios \cite{radford2021learning,koutlis2023memefier}.
Since the test set labels are not publicly available, we follow existing studies to evaluate models on the development set.
The statistics of the datasets used in the experiments are reported in Table \ref{tab:dataset_statistics}.
}

\subsection{Baselines}
\label{sec:baselines}

To validate the effectiveness of our proposed approach, we compare it against the following three baseline models.
The first (``\textbf{LLM}'') directly leverages a pre-trained multi-modal LLM to encode the input image and text and
does not include any explicit representation operations, especially the mechanisms to capture cross-modal contrast information.
The second baseline, named ``\textbf{LLM + LMR}'', adds the standard LMR on top of LLM, 
where it decomposes each modality's representation into a low-rank component and a sparse component, which does not utilize the shared (common) and modality-specific information or the attention mechanisms.
The third baseline, named ``\textbf{LLM + RD}'', adopts the same visual and textual encoding and representation decomposition procedure as our proposed approach, whereas the attention mechanism is not used with all components aggregated using the same weights.

\begin{table}[t]
    \centering
    \scalebox{1.0}{
    \begin{tabular}{ L{4cm} | R{1.2cm} R{1.2cm} }
        \toprule
        & \multicolumn{2}{c}{\textbf{Twitter-15}} \\
        \addlinespace[0.05cm]
        \cline{2-3}
        \addlinespace[0.05cm]
        & \multicolumn{1}{c}{\textbf{Acc}} & \multicolumn{1}{c}{\textbf{F1}} \\
        \toprule
        \citet{xiao2023cross} 
        & 79.9\phantom{0} & 75.3\phantom{0} \\
        \citet{zhou-etal-2023-aom} 
        & 80.2\phantom{0} & 75.9\phantom{0} \\
        \citet{lawan2024dualkanbaformer} 
        & 78.71 & 75.53 \\
        \citet{huang2024utilizing} 
        & - & 71.0\phantom{0} \\
        \citet{yang2024large} 
        & 53.85 & - \\
        \citet{yang2024empirical} 
        & 62.35 & - \\
        \citet{fan2024dual} 
        & 80.71 & \textbf{77.15} \\
        \midrule
        Ours (LLaVA (13B))
        & \textbf{81.26} & 76.52 \\
        \bottomrule
    \end{tabular}
    }
    \caption{
    Comparison of our approach with existing studies on the MABSA task with multi-modal inputs.}
    \vspace{-0.5em}
    \label{tab:sota-mabsa}
\end{table}

\begin{table}[t]
\centering
\scalebox{0.9}{
\begin{tabular}{l | cc}
\toprule
& \multicolumn{2}{c}{\textbf{MSED}} \\
        \addlinespace[0.05cm]
        \cline{2-3}
        \addlinespace[0.05cm]
& \textbf{Acc} & \textbf{F1}\\
\midrule
BERT+ResNet \cite{jia-etal-2022-beyond} & - & 82.42\\
ViLT \cite{aziz2023mmtf} & - & 80.81\\
M3GAT \cite{zhang2023m3gat} & - & 81.97\\
\midrule
Ours (LLaVA (13B)) & \textbf{84.33} & \textbf{83.36}\\
\bottomrule
\end{tabular}
}
\caption{
Performance comparison of our approach and existing studies on the test set of MSED for MEA.
}
\label{table:sota-mea}
\vspace{-0.3cm}
\end{table}

\begin{table}[t]
    \centering
        \centering
        \scalebox{0.93}{
        \begin{tabular}{ l | c c  }
            \toprule
            & \multicolumn{2}{c}{\textbf{HMC}}
            \\
                    \addlinespace[0.05cm]
        \cline{2-3}
        \addlinespace[0.05cm]
            & \textbf{ACC} & \textbf{AUROC}
            \\
            \midrule
            
            ViLBERT CC
            & 65.90 & 74.52 \\

            Visual BERT COCO
            & 69.47 & 75.44 \\

            \citet{cao2023prompting}
            & 72.98 & 82.45
            \\

            \citet{koutlis2023memefier}
            & 73.60 & 80.10
            \\

            $\triangle$\citet{liu2023llava}
            & 76.20 & 84.57
            \\

            \citet{mei-etal-2024-improving}
            & \textbf{78.30} & \textbf{86.70}
            \\
            
            \midrule
            Ours (LLaVA (13B))
            & {77.00}
            & {85.36}
            \\
            \bottomrule
        \end{tabular}
        }
        \vspace{-0.1cm}
        \caption{
        \label{tab:sota-hmc}
        Comparing our approach with existing studies on the development set of HMC. 
        ``$\triangle$'' marks the results of our own runs of existing multi-modal LLMs.
        }
        \vspace{-0.3cm}
\end{table}

\subsection{Implementation Details}

Consider pre-trained embeddings or models have achieved remarkable performance on many tasks \cite{mikolov2013efficient,song-etal-2017-learning,song2018joint,song2018complementary,zhang-etal-2019-incorporating,peters-etal-2018-deep,han-etal-2018-hyperdoc2vec,devlin-etal-2019-bert,song2021zen,touvron2023llama,taori2023alpaca}, 
For the visual and textual encoding and alignment, we utilize the CLIP model \cite{radford2021learning} to obtain cross-modal representations.
Particularly, the input images are all resized to $224\times224$ to satisfy the input requirements of the vision models.
For the representation decomposition module, the penalty parameter is set to $\mu=10$, and the balancing parameter is set to $\lambda=1$. 
The decomposition process runs for 3000 iterations to ensure convergence of the low-rank and sparse components.\footnote{\textcolor{black}{We only run this process once and use the decomposed components as the hard input of our approach, which does not lead to extra training time. For reference, the average running time for each image-text pair is 1.3 seconds running on Intel(R) Xeon(R) Platinum 8375C CPU.}}
\textcolor{black}{
We employ Qwen2-VL (2B) \cite{wang2024qwen2}, LLaVA-1.5 (7B) \cite{liu2023llava}, and LLaVA-1.5 (13B) as the foundation LLMs\footnote{\textcolor{black}{We download Qwen2-VL (2B), LLaVA-1.5 (7B), and LLaVA-1.5 (13B) from \url{https://huggingface.co/Qwen/Qwen2-VL-2B-Instruct}, , \url{https://huggingface.co/llava-hf/llava-1.5-7b-hf}, and \url{https://huggingface.co/llava-hf/llava-1.5-13b-hf}, respectively.}} for affective computing. 
For Qwen2-VL (2B), its vision encoder consists of 32 layers of Transformer with 1,280-dimensional hidden states, and its LLM is configured with a hidden state size of 1,536 and 28 transformer layers.
For LLaVA-1.5 (7B), the vision encoder contains 24 layers of Transformers with 1,024-dimensional hidden vectors, and the LLM is a 32-layer Transformer with 4,096-dimensional hidden vectors.
For LLaVA-1.5 (13B), it uses the same vision encoder as the 7B version, and the LLM employs a Transformer with 40 layers and 5,120-dimensional hidden vectors.
}
For evaluation, we use accuracy and macro F1 score as the metrics for both MABSA and MEA tasks. 
For the HMD task, we use accuracy and AUROC.

\section{Results and Analysis}

\subsection{Overall Results}

Table \ref{tab:overall_results} presents the overall performance of our approach and the baselines across all evaluated tasks, with several observations.
First, compared to the vanilla LLM baseline, our approach consistently achieves higher performance \textcolor{black}{with different types of LLMs},
which demonstrates that explicitly modeling cross-modal relations via representation decomposition is beneficial for capturing subtle differences between modalities.
Second, when compared with the ``LLM + LMR'' baseline, our approach shows performance gains for the reason that the explicit separation of shared and modality-specific components enables our approach to better handle conflicting signals, leading to more robust and discriminative representations.
Third, our approach further outperforms the ``LLM + RD'' baseline. 
The integration of the attention mechanism allows for adaptive fusion of the decomposed components, thereby emphasizing the most relevant information for task label prediction.\footnote{\textcolor{black}{We randomly sample 10 cases from each dataset (30 cases in total) where our model is able to make correct predictions, yet the vanilla LLM cannot, and manually check whether there are modality conflicts. Among the 30 cases, 19 of them are identified to have modality conflicts, which further confirms the effectiveness of our approach.}}

We further compare our approach \textcolor{black}{(using LLaVA-1.5 (13B))} with existing studies and report the results for MABSA, MEA, and HMD in Table \ref{tab:sota-mabsa}, Table \ref{table:sota-mea}, Table \ref{tab:sota-hmc}, respectively, where experiment results show that our approach is versatile and outperforms existing studies on multiple affective computing tasks, where state-of-the-art results are observed on Twitter-15 and MSED and relatively best performance is obtained on HMC.
This observation is attributed to the explicit decomposition process of representations into shared and modality-specific components, allowing the model to capture commonalities while exploiting contrastive evidence that conventional fusion techniques often overlook.
Particularly, for HMD, emphasizing the contrast between visual and textual inputs enables the model to pinpoint subtle discrepancies indicative of hateful content, especially when either modality alone proves insufficient.
Similarly, in tasks such as MABSA and MEA, the integration of an attention mechanism facilitates adaptive fusion by dynamically weighting the most informative components, thereby reducing noise and enhancing the discriminative power of overall representations.
\textcolor{black}{
In addition, compared to the study \citet{mei-etal-2024-improving} utilizing extra resources, our approach still shows comparable performance, which further illustrates its effectiveness.
}

\begin{figure}[t]
    \centering
    \includegraphics[width=1\linewidth, trim=0 60 0 0]{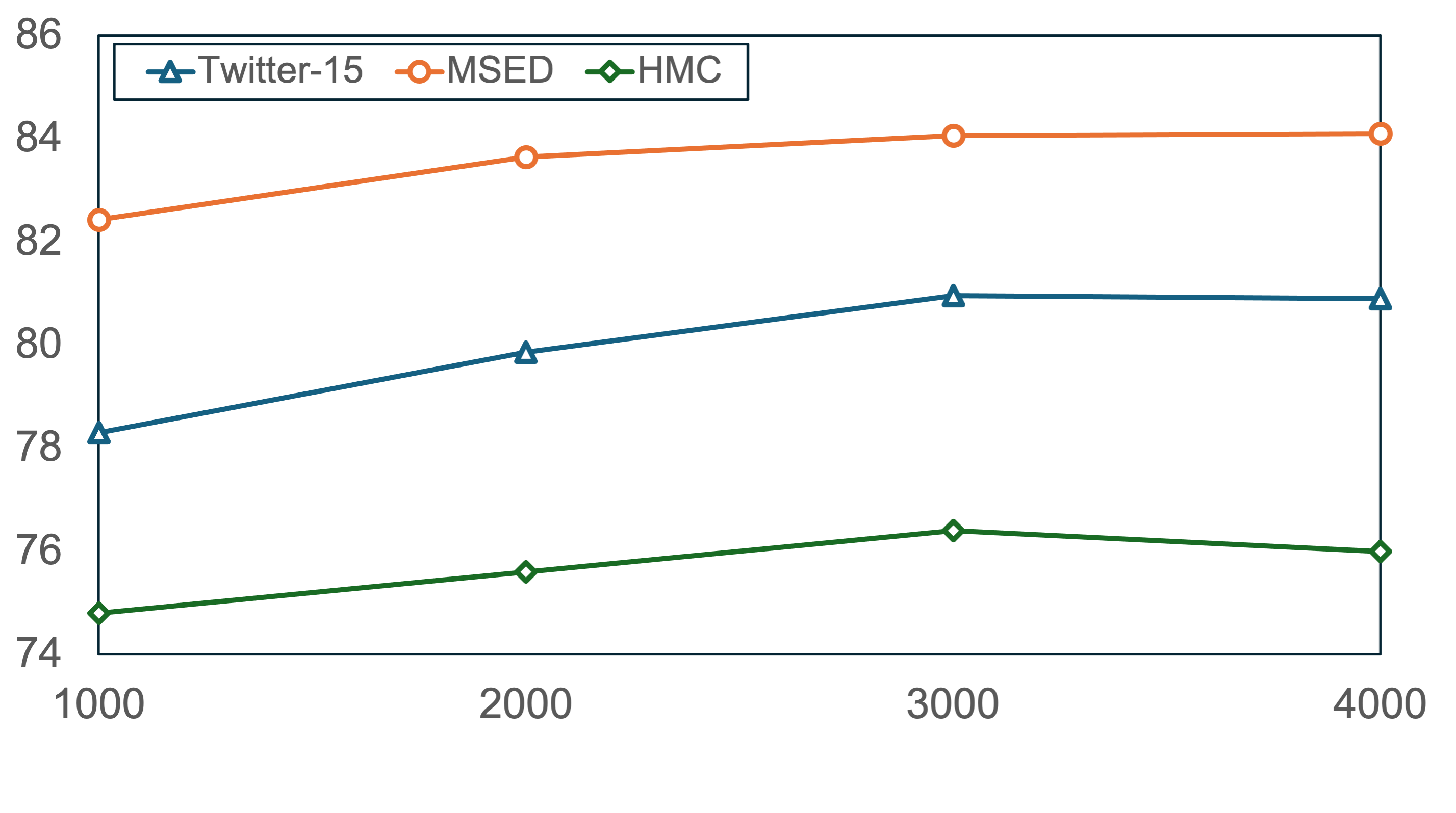}
    \caption{
    Accuracy curves of our approach \textcolor{black}{(with Qwen2-VL (2B))} with respect to different numbers of running iterations (i.e., 1000, 2000, 3000, and 4000) to perform representation decomposition.}
    \label{fig:iteration_effect}
    \vspace{-0.5em}
\end{figure}

\subsection{Effect of Representation Decomposition}

Since representation decomposition is largely affected by how the shared LMR performs,
in this analysis, we investigate how different numbers of iterations in the representation decomposition process affect model performance.
Specifically, we try iteration 1000, 2000, 3000, and 4000, respectively, where fewer iterations (e.g., 1000) means that the shared component \(\mathbf{L}\) captures less common information, causing much of the original features to remain in the modality-specific sparse components (\(\mathbf{S}_I\) and \(\mathbf{S}_T\)).
With the iteration number increasing, the separation between the shared and sparse components becomes clearer, which allows the model to better isolate common emotional cues from modality-specific nuances.
Figure \ref{fig:iteration_effect} presents the accuracy of our approach \textcolor{black}{(using Qwen2-VL (2B))} with respect to the number of iterations. 
The results indicate that performance improves when increasing the iterations from 1000 to 3000, suggesting that the decomposition is effectively refining the shared representation during this range. 
Beyond 3000 iterations, the performance tends to be stable and the gains are relatively small, implying that the decomposition process has largely converged and that further iterations do not provide more information.
This behavior validates our shared LMR design and confirms that a moderate number of iterations (around 3000) is sufficient for achieving an optimal separation of the shared and modality-specific components.

\begin{table}[t]
    \centering
    \scalebox{0.9}{
    \begin{tabular}{l|ccc}
        \toprule
        & \multicolumn{1}{c}{\textbf{Twitter-15}} & \multicolumn{1}{c}{\textbf{MSED}} & \multicolumn{1}{c}{\textbf{HMC}} \\
         \cline{2-4}
         & \textbf{F1} 
         & \textbf{F1} 
         & \textbf{AUROC} 
         \\
        \midrule
        Full Model 
        & 76.30 
        & 83.15 
        & 85.04 \\
        \midrule
        - Shared (\(L\))     
        & 74.23 
        & 85.67 
        & 83.42 \\
        - Sparse (\(S\))     
        & 75.14 
        & 86.58 
        & 84.74 \\
        \midrule
        Only $S_I$ and $S_T$ 
        & 72.54 
        & 84.48 
        & 82.74 \\
        Only $L$ 
        & 72.84 
        & 84.79 
        & 83.85 \\
        \bottomrule
    \end{tabular}
    }
    \vspace{-0.1cm}
    \caption{
    \textcolor{black}{Ablation study results on benchmark datasets. The full model consistently outperforms others that exclude either the shared component (\(L\)) or the modality-specific sparse components (\(S_I\) and \(S_T\)), or use the modality-specific or the shared components only.
    The LLM used in this experiment is Qwen2-VL (2B).}}
    \label{tab:ablation}
    \vspace{-0.5cm}
\end{table}

\begin{figure*}[t]
    \centering
    \includegraphics[width=0.97\linewidth, trim=0 15 0 0]{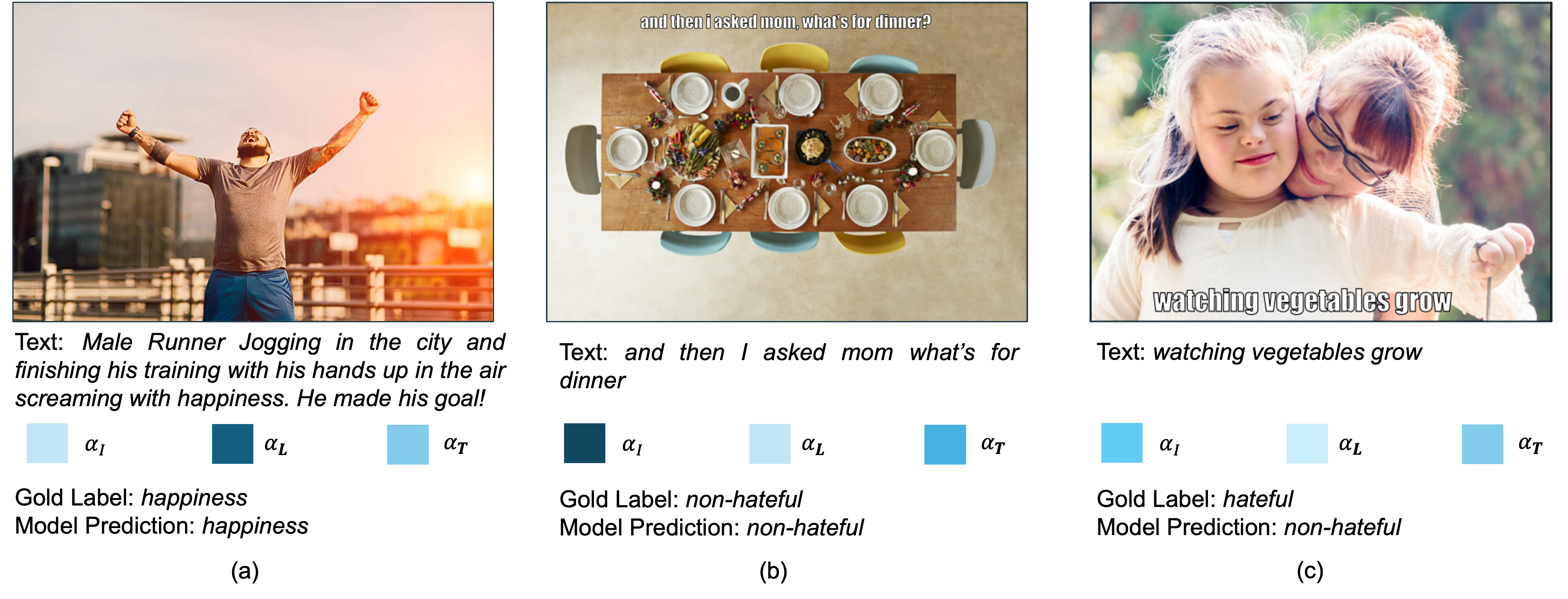}
    \caption{
    Demonstration of \textcolor{black}{three} examples with the image and text inputs to show the effect of the attentive soft prompting approach over the shared and model-specific representations for different image-text relations.
    $\alpha_L$, $\alpha_I$, and $\alpha_T$ denote the attention weights for the shared representation, the image representation, and the text representation, respectively, where darker colors refer to higher weight values.
    \textcolor{black}{
    For the first two cases, our model is able to correctly predict the labels with appropriate weights for different components.
    For the last case, our model is able to distinguish the conflict in the image and text, yet is unable to distinguish the hateful information.}
    }
    \label{fig:case-study}
    \vspace{-0.2cm}
\end{figure*}

\subsection{Ablation Study}

To verify the contributions of each component resulted from our representation decomposition, we conduct ablation experiments \textcolor{black}{using Qwen2.5-VL (2B)} on all three datasets (Twitter-15, MSED, and HMC).
Particularly, we investigate \textcolor{black}{four} variants: (1) removing the shared low-rank component $\mathbf{L}$ so that only the modality-specific sparse components (\(\mathbf{S}_I\) and \(\mathbf{S}_T\)) are used; (2) removing the sparse components so that the model relies solely on the shared component \(\mathbf{L}\); \textcolor{black}{(3) only using the sparse components without the original image $I$ and text $T$; and (4) only using the low-rank component without $I$ and $T$.}
As shown in Table \ref{tab:ablation}, the full model achieves the best performance across all evaluation metrics.
The ablation of the shared component $\mathbf{L}$ leads to a significant drop in performance, indicating that capturing the common emotional evidence is critical for accurate affective interpretation.
Similarly, when the modality-specific information (i.e., $\mathbf{S}_I$ and $\mathbf{S}_T$) is removed, the model’s capability to capture contrastive signals is also hurt, which is especially evident in the HMD task, because in this task there are cases that require the understanding of the contrasts between modalities to correctly identify whether the content is hateful.
\textcolor{black}{
Moreover, a further drop in performance is observed if we further ablate the original image and text, where only the sparse components or the low-rank components are used.
The observation demonstrates that the original image and text still provide essential information that is helpful for affective computing.}

\subsection{Case Study}

We present a case study that visualizes how the attention mechanism dynamically adjusts the weights of the decomposed components across different cases.
Figure \ref{fig:case-study} displays the attention distribution (i.e., $\alpha_L$, $\alpha_I$, and $\alpha_T$) for the shared (low-rank) and modality-specific (sparse) matrices when processing examples from diverse affective computing scenarios, where darker color stands for higher weights when our model's predictions match the gold standard labels.
\textcolor{black}{
As illustrated in Figure \ref{fig:case-study}(a), when the modalities are highly consistent (i.e., both convey a positive sentiment), the attention weight on the shared component is dominant, emphasizing the common emotional content (i.e., \textit{happiness}).}
In contrast, as presented in Figure \ref{fig:case-study}(b), for cases where there is a marked contrast between the modalities, the attention mechanism allocates higher weights to the modality-specific components, thereby capturing the crucial conflicting evidence.
This adaptive behavior demonstrates the capability of our approach to focus on the most informative aspects of the input, so that meeting the specific requirements of each task.
\textcolor{black}{
Moreover, for Figure \ref{fig:case-study}(c), our model is able to distinguish the conflict in the image and text and thus assigns high weights to the sparse components.
However, for a model, it is difficult to directly combine the image’s meaning of ``\textit{a mother watching her daughter grow}'' with the phrase ``\textit{watching vegetables grow}'', and infer the hidden analogy that the daughter is a vegetable, making it hard to judge whether it constitutes hate speech.}

\section{Related Work}

Multi-modal affective computing focuses on recognizing and interpreting human emotions by integrating signals from diverse modalities such as images and text, including typical tasks such as MABSA, MEA, and HMD, etc.
Conventional studies primarily rely on unimodal approaches, which use either image or text information to perform affective computing \cite{ren2015faster,keswani-etal-2020-iitk-semeval,li-etal-2021-dual-graph,mei-etal-2024-improving}.
Subsequent research introduces multi-modal techniques, merging image and text encoders for enhanced affective computing, employing advanced encoders (such as ViLBERT \cite{lu2019vilbert}, CLIP \cite{radford2021learning}, and FLAVA \cite{singh2022flava}) to analyze both image and text features,
and then integrating these multi-modal features through specific modules, such as vector concatenation and attentions \cite{giri2021approach,kirk2021memes,kumari-etal-2021-co,tsimpoukelli2021multimodal,gu2021targeted,huang2021attention,shang2021aomd,ouaari2022multimodal,wang2022cross,hee2023decoding,qu2023on,huang2023dominant,hu2024vision,wang2024aspect}.
To further enhance affective computing, model ensemble \cite{guo-etal-2020-guoym,lippe2020multimodal,sandulescu2020detecting}, multi-task learning \cite{vlad-etal-2020-upb,zhang2023multitask}, 
additional resources (e.g., extra training data, image segmentation results, keywords, and syntactic information) \cite{velioglu2020detecting,zhou2021multimodal,wang2023dual}, contrastive learning \cite{liang2022mind}, and language model prompting \cite{cao2023prompting} are employed to improve the ability to capture multi-modal features.
More recently, multi-modal LLMs \cite{liu2023visual,bai2023qwen} integrate visual and textual processing within the unified transformer framework using prompt-based mechanisms to implicitly align inputs from multiple modalities. 
Our approach differs from existing studies by explicitly decomposing the aligned multi-modal representations into shared and modality-specific components using a low-rank matrix recovery framework, which is then dynamically fused via an attention mechanism to instruct the LLM to perform affective computing tasks.
To the best of our knowledge, there is no related work using similar approaches that is effective for different affective computing tasks.

\section{Conclusion}

In this paper, we propose a novel approach for multi-modal affective computing by explicitly learning and leveraging the alignments and contrasts across modalities.
Our approach decomposes aligned visual and textual representations into shared (modality-invariant) and modality-specific components, enabling the model to capture both common emotional content and unique, expressive evidence.
In enhancing the decomposed representations, we further improve our approach with attentive soft prompting to dynamically utilize these components to instruct the LLM to perform affective computing tasks.
Extensive experiments on MABSA, MEA, and HMD demonstrate that our approach consistently outperforms strong baselines and existing studies, which confirms the effectiveness of explicit representation decomposition and adaptive fusion, as well as the generalization ability in tackling the challenges of affective computing.

\bibliography{custom,ref}

\end{document}